\documentclass[10pt,twocolumn,letterpaper]{article}

\usepackage{cvpr}              

\usepackage{graphicx}
\usepackage{amsmath}
\usepackage{amssymb}
\usepackage{booktabs}
\usepackage{booktabs}
\usepackage{multirow}
\usepackage[pagebackref,breaklinks,colorlinks]{hyperref}

\usepackage[capitalize]{cleveref}
\crefname{section}{Sec.}{Secs.}
\Crefname{section}{Section}{Sections}
\Crefname{table}{Table}{Tables}
\crefname{table}{Tab.}{Tabs.}

\def\cvprPaperID{2825} 
\def\confName{CVPR}
\def\confYear{2022}

\begin{document}


\title{CLIP-TD: CLIP Targeted Distillation for Vision-Language Tasks }

\author{Zhecan Wang\thanks{Equal Contribution. Correspondence to: Zhecan Wang $\langle$olinzhecanwang@gmail.com$\rangle$, Noel Codella $\langle$ncodella@microsoft.com$\rangle$.}   $^1$
  \and Noel Codella$^*$$^2$
  \and Yen-Chun Chen$^2$
  \and Luowei Zhou$^2$
  \and Xiyang Dai$^2$
  \and Bin Xiao$^2$
  \and Jianwei Yang$^2$
  \and Haoxuan You$^1$
  \and Shih-Fu Chang$^1$
  \and Lu Yuan$^2$}
\maketitle


\footnotetext[1]{Columbia University}
\footnotetext[2]{Microsoft Research}

\begin{abstract}
Contrastive language-image pretraining (CLIP) links vision and language modalities into a unified embedding space, yielding tremendous potential for vision-language (VL) tasks. While early concurrent works have begun to study this potential on a subset of tasks, important questions remain: 1) What is the benefit of CLIP on unstudied VL tasks? 2) Does CLIP provide benefit in low-shot or domain shifted scenarios? 3) Can CLIP improve existing approaches without impacting inference or pretraining complexity? In this work, we seek to answer these questions through two key contributions. First, we introduce an evaluation protocol that includes  Visual Commonsense Reasoning (VCR), Visual Entailment (SNLI-VE), and Visual Question Answering (VQA), across a variety of data availability constraints and conditions of domain shift. Second, we propose an approach, named CLIP Targeted Distillation (CLIP-TD), to intelligently distill knowledge from CLIP into existing architectures using a dynamically weighted objective applied to adaptively selected tokens per instance. Experiments demonstrate that our proposed CLIP-TD leads to exceptional gains in the low-shot (up to 51.9\%) and domain-shifted (up to 71.3\%) conditions of VCR, while simultaneously improving performance under standard fully-supervised conditions (up to 2\%), achieving state-of-art performance on VCR compared to other single models that are pretrained with image-text data only. On SNLI-VE, CLIP-TD produces significant gains in low-shot conditions (up to 6.6\%) as well as fully supervised (up to 3\%). On VQA, CLIP-TD provides improvement in low-shot (up to 9\%), and in fully-supervised (up to 1.3\%). Finally,  CLIP-TD outperforms concurrent works utilizing CLIP for finetuning, as well as baseline naive distillation approaches. Code will be made available.

\end{abstract}

\vspace{-6mm}
\section{Introduction}
\label{sec:intro}

\begin{figure}[t!]
\centering
\includegraphics[width=1\linewidth]{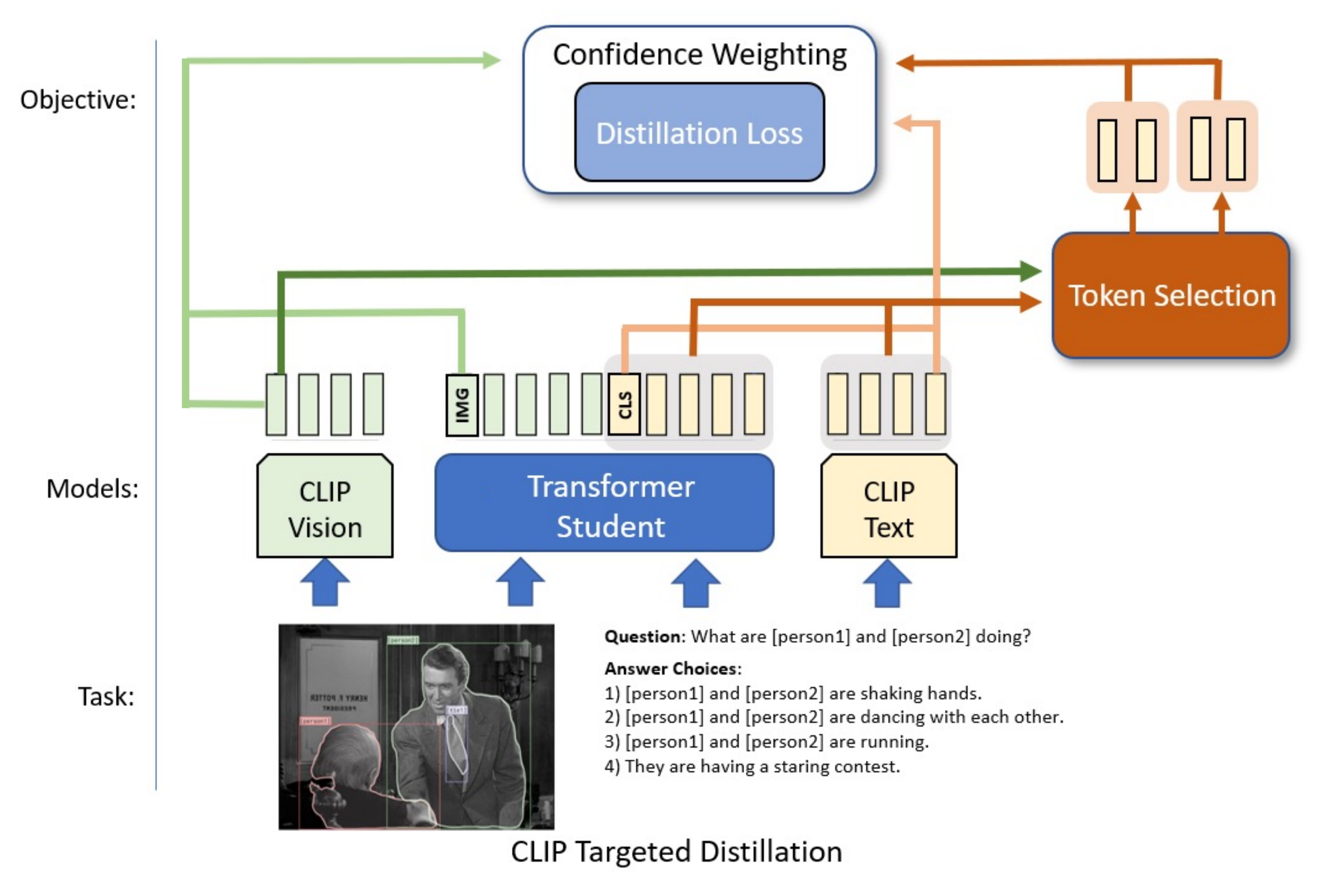}

\caption{The proposed CLIP-TD targeted distillation. For a given task with an established base model as the student, a token selection method chooses tokens to distill from a CLIP teacher to the student. Then confidence weighting determines the weight for this distillation instance, based on confidence of the teacher and the student. We perform thorough ablation to understand contributions of each component, and compare to naive distillation, which removes both confidence weighting and token selection.}
\vspace{-5mm}
\label{fig:diagram2}
\end{figure}

Visual Commonsense Reasoning (VCR) \cite{zellers2019vcr}, Stanford Natural Language Inference Visual Entailment (SNLI-VE) \cite{xie2019visual}, and Visual Question Answering (VQA) \cite{VQA}, are all vision-language (VL) tasks requiring solutions that effectively bridge the two modalities of vision and language, with mechanisms of logic and prior knowledge serving as the link between them. For example, in VCR, logical reasoning capability and common knowledge is useful in answering questions about images, especially when the visual information is ambiguous, rendering the task of connecting questions with their correct answers difficult.

Considering this, most recent works that have achieved state-of-art on these tasks tend to capture logic and prior knowledge with various cross-modal transformer architectures that jointly model both modalities in a unified architecture ~\cite{chen2020uniter,Su2020VL-BERT,gan2020large,huang2021seeing,huang2020pixel,lxmert}. These models are typically pretrained on large vision-language datasets composed of paired images and text, and optimized via a combination of supervised and self-supervised objective functions. However, these datasets, while large by one definition, are still limited by other definitions: they typically do not exceed about 10M data points in size.

Concurrently, a separate line of research has been examining training modality specific encoders combined with a contrastive language-image pretraining (CLIP) objective to align both modalities into a shared embedding space, using an order of magnitude more data (400M to 1B image-text pairs) ~\cite{clip,align}. These pretrained models are then studied in a variety of downstream tasks, which have included classification and object detection ~\cite{clip,align,googleod}.

While the pretrained knowledge captured by CLIP models has the potential to augment existing VL approaches, comparatively few works have examined what implications these large models might have on vision-language tasks, especially regarding more challenging and highly semantic tasks such as VCR. One concurrent work \cite{shengshen} studies the effectiveness of incorporating CLIP's vision branch to VL tasks such as VQA ~\cite{VQA}, COCO captioning ~\cite{cococaptions}, and SNLI-VE ~\cite{shengshen}. This work demonstrated noticeable gains in performance over baselines within these benchmarks. However, the following questions remain: 1) What is the benefit of CLIP on unstudied VL tasks, such as VCR? 2) What benefit does CLIP provide in low-shot and domain shifted scenarios, which better reflect conditions of learning in new and unique domains? 3) Can CLIP improve existing approaches without impacting inference or pretraining complexity? In other words, can CLIP give us a ``free lunch''?

In this work, we seek to answer these remaining questions through the following two contributions:

{\bf 1) Comprehensive VL Protocol:} We propose a new vision-language evaluation protocol involving {\em highly-semantic} VL tasks (VCR, SNLI-VE) and VQA, where evaluation is conducted under a variety of data availability constraints, including zero-shot, low-shot, semi-supervised, and fully-supervised settings. For VCR, we additionally use established and new ways of perturbing the characteristics of the evaluation set to measure model performance under domain-shift. To our knowledge, we are the {\em first} to perform such comprehensive analysis, especially zero-shot and low-shot analysis on VCR and SNLI-VE.

{\bf 2) CLIP Targeted Distillation (CLIP-TD): } We propose an approach, named CLIP-TD, to intelligently distill knowledge from both CLIP's vision and language branches into existing architectures for VL tasks. The key idea is that knowledge distillation dynamically adapts {\em per instance} by selecting tokens and weighting distillation according to teacher confidence values. This method does not require redoing tedious and computationally expensive pretraining steps, maintains inference complexity of the student model, and is agnostic to the base student network. We compare this method to a variety of other baselines, and demonstrate superior performance, achieving state-of-art on VCR Q2A for single models pretrained with image-text data only.

In the following sections, we more thoroughly review the related work. Then we introduce our proposed approach, as well as the baselines, followed by a thorough description of our evaluation protocol. Finally, we present the results.

\vspace{-2mm}
\section{Related Work}


\textbf{Vision-Language Pretraining: }  Joint vision-language pretraining has been an effective approach to improve performance on VL tasks. Many works, such as LX-MERT \cite{lxmert}, VL-BERT \cite{Su2020VL-BERT}, UNITER \cite{chen2020uniter}, VILLA \cite{gan2020large}, and others \cite{huang2021seeing,huang2020pixel,zhou2020unified,kim2021vilt,yu2020ernie,li2021align} have leveraged this approach. All of these works utilized datasets on the order of 10M samples for various pretraining objectives.

Following this, works such as CLIP \cite{clip}, ALIGN \cite{ jia2021scaling}, and SimVLM \cite{wang2021simvlm} extended pretraining to orders of magnitude more image-text data (400 million +), demonstrating utility on a wide range of downstream tasks. CLIP showed strong performance for zero-shot and linear adapter classification tasks. ALIGN demonstrated strong performance on VL tasks, classification, and captioning. SimVLM has focused on VL tasks such as VQA, SNLI-VE, and captioning. SimVLM also claims to evaluate zero-shot analysis on SNLI-VE, but pretrains on in-domain text data. To our knowledge, no prior contrastive pretraining works have studied impact on generalization capability by assessing performance on highly-semantic tasks, such as VCR, under both low-shot and domain shifted scenarios, which are more reflective of the challenges encountered in practice. The closest concurrent work, CLIP-ViL \cite{shen2021clip} proposed applying additional pretraining and fine-tuning to their base CLIP model  for tasks including image-text retrieval, SNLI-VE, and VQA. However, this work required additional pretraining and a customized architecture. Currently, there is a lack of general frameworks for leveraging large scale pretrained models for a variety of downstream VL tasks, especially while controlling inference complexity.

{\em In this work}, we address several of these mentioned gaps. First, we study the impact of CLIP on VL tasks including VCR, SNLI-VE, and VQA. Second, we assess  the impact of CLIP in true zero-shot, low-shot, and domain shifted scenarios. Third, we propose an efficient distillation approach to leverage CLIP in a way that doesn't require additional pretraining and retains inference complexity.



\textbf{Generalization of Question Answering: }Visual question answering \cite{zellers2019vcr, VQA} is among the most challenging of VL tasks, due to large variation between samples and distribution discrepancy between training and testing datasets. This results in difficulty for models to perform well under zero-shot, low-shot, and domain shifted settings. Despite high accuracy recent models achieve \cite{su2019vl, gan2020large, li2020oscar, zhang2021vinvl, zellersluhessel2021merlot}, other works have begun to uncover the tendency of models to leverage spurious shortcut signals, or memorization of mapping distributions \cite{sen2020models, jiang2019avoiding, dancette2021beyond, kovaleva2019revealing}. While prior works have begun to explore question answering tasks under zero-shot \cite{teney2016zero, noh2019transfer}, low-shot \cite{chada2021fewshotqa, brown2020language} and domain shifted \cite{dancette2021beyond, jiang2019avoiding, goyal2017making, zhang2016yin, agrawal2018don, shah2019cycle, ramakrishnan2018overcoming} settings, they are mostly limited within text-only question answering and low-level visual question answering tasks (\textit{i.e.} VQA \cite{VQA}). To the best of our knowledge, none of the prior works have explored true zero-shot and low-shot settings in highly complex visual question answering datasets such as VCR or SNLI-VE, and only one prior work \cite{debias} contributed to VCR with domain shift.

{\em In this work}, we conduct a thorough evaluation with many existing top-performing models on zero-shot, low-shot, and domain-shifted settings. We further proposed an novel domain-shifted evaluation protocol on VCR.

\textbf{Knowledge Distillation: }In conventional knowledge distillation \cite{liu2020adaptive, yang2020knowledge, wang2021knowledge, mun2018learning, googleod, cho2021dealing, sanh2019distilbert, jiao2019tinybert, kim2016sequence, sun2019patient}, a larger model serves as the teacher to a smaller student model. In self-distillation scenarios \cite{furlanello2018born, cheng2021data, li2021align}, the size of the teacher model is typically comparable to student model.  The goal is usually to obtain a computationally lighter and more efficient framework, but still maintain similar or even higher accuracy. But no prior works have addressed the feasibility of distilling smaller teacher models to much larger student models. Further, in most of these scenarios, the components of the networks leveraged for the distillation loss is fixed, and the weight of the distillation loss is also fixed. One prior work first introduced the idea of dynamically adjusting the distillation loss weight on a per instance level \cite{furlanello2018born}.
{\em In this work}, we leverage the idea of dynamic distillation weights, but we go one step further: we dynamically adjust not only the {\em distillation weight}, but {\em which parts of the network are distilled} as well. We also address the feasibility of distilling from a small teacher model into a larger student with our proposed CLIP-TD method.




\vspace{-1mm}
\section{Methods}
\vspace{-1mm}
In the following, we discuss the naive baseline distillation approach, our proposed CLIP Targeted Distillation approach, and the student models. We also include other baselines of direct CLIP training/fine-tuning. For training parameters, please refer to the supplement.



\vspace{-1mm}
\subsection{Knowledge Distillation}

\subsubsection{Naive Knowledge Distillation}
\vspace{-1mm}
Naive distillation is equivalent to CLIP-TD when both confidence weighting and token selection are omitted (Fig. \ref{fig:diagram2}). The CLIP visual $cls$ token and student model $img$ token, in addition to the CLIP text $eos$ token and the student text $cls$ tokens, are compared via L1 measure. The final loss is the weighted distillation loss $L_{d}$ summed with the original task loss $L_{t}$ for any specific downstream task. Formally,
\begin{equation}
\begin{aligned}
L_{f i n a l} &=L_{t}+w \cdot\left(L_{d}\right)
\end{aligned}
\label{eq:vl}
\end{equation}
\vspace{-1mm}

\noindent where the distillation loss $L_{d}$ is a  sum of the distillation losses between the two modalities:
\begin{equation}
\begin{aligned}
L_{d} &=L_{d,v}+L_{d,t}
\end{aligned}
\end{equation}

\noindent where $v$ and $t$ refer to vision and text branches, respectively.
\begin{equation}
\begin{aligned}
L_{d,v} &=\left\|f_{c, v,cls}\left(i_{j}\right)-f_{b,v,img}\left(\partial\left(i_{j}\right)\right)\right\|_{1} \\
\end{aligned}
\end{equation}

\noindent where $f_{c,v,cls}$ and $f_{b,v,img}$ refer to the visual $cls$ token of the CLIP vision encoder, and the base model $img$ token. $\partial$ represents the backbone detection network. $i_{j}$ refers to the image input for instance $j$. For the text modality, we have
\begin{equation}
\begin{aligned}
L_{d,t} &=\left\|f_{c,t,eos}\left(t_{j}\right)-f_{b,cls}\left(t_{j}\right)\right\|_{1}
\end{aligned}
\end{equation}

\noindent where $f_{c,t,eos}$ refers to the $eos$ token of the CLIP text model, and $f_{b,cls}$ refers to the $cls$ token of the base model. $t_{j}$ refers to the text input for instance $j$.







\vspace{-4mm}
\subsubsection{CLIP Targeted Distillation (CLIP-TD) }
\vspace{-1mm}
The CLIP Targeted Distillation method consists of the following 3 components: Token Selective Distillation using an unsupervised language prior, Confidence Weighted Distillation, and Adaptive Finetuning with Contrastive Knowledge.

{\bf \noindent Token Selective (TS) Distillation with Prior:} Typically, distillation methods \cite{sanh2019distilbert, jiao2019tinybert, kim2016sequence, sun2019patient} distill knowledge from teacher to student using fixed components of a network. However, the most semantically relevant tokens can change {\em per instance}. Without a dynamic process to determine which components of a network should be distilled, the student model may have a higher risk of learning spurious signals from trivial tokens. Addressing this, we present a hybrid method of performing token selection for distillation.

Given a text sequence $t_{j}=\left\{w_{0} \ldots w_{z}\right\}$, where $z$ is the length of the sequence, we apply a pretrained Token Selection Module (TSM) $f_{tsm}\left(i_{j},t_{j}\right)$ to discriminate the semantically meaningful tokens. TSM generates a score, $s_{l}$ for each token $w_{l}$, obtaining a distribution $S_{j}=\left\{s_{0} \ldots s_{z}\right\} = f_{tsm}\left(i_{j}, \left\{w_{0} \ldots w_{z}\right\}\right)$. In our implementation of this selective module, two sets of weights are computed via two different approaches, and summed together:

\vspace{-2mm}
\begin{equation}
\begin{gathered}
S_{j}=\frac{S_{vr}}{|S_{vr}|_{1}} + \frac{S_{si}}{|S_{si}|_{1}} \\
\end{gathered}
\end{equation}

\noindent where $S_{vr} = \left\{s_{vr0} \ldots s_{vrz}\right\} = \left\{cos\left<f_{c, v}\left(i_{j}\right), f_{c,t}\left(w_{l}\right)\right>\right\}$ represents a scoring between the CLIP image embedding and each of its text token embeddings. In this manner, $S_{vr}$ is the score measuring the visual relevance of each token $w_{l}, l \in[0, \ldots, z]$. Then, $S_{si} = f_{ke}\left(t_{j}\right)$, where $S_{si}$ represents
the semantic and syntactic importance of each token related to the context, $t_{j}$, using a keyword extractor. In practice, we apply a pre-trained keyword extractor\cite{yake}, $f_{ke}$ with $n$-grams:





\vspace{-1mm}

\begin{equation}
t_{j}^{\prime}=f_{argmax}\left(S_{j}, t_{j}, m \right)
\end{equation}


\noindent $f_{argmax}$ is a function that would select $m$ tokens with the highest scores, $t_{j}^{\prime}$, where $\left|t_{j}{ }^{\prime}\right| = m$. The corresponding features of both CLIP and base model would be compared with an L1 measure to calculate their difference:

\vspace{-2mm}
\begin{equation}
L_{d t}{ }^{\prime}=\left\| f_{c, t}\left( t_{j}{ }^{\prime}\right)-  f_{b}\left(t_{j}{ }^{\prime}\right)\right\|_{1}
\end{equation}

\noindent finally, the outputted loss, $L_{d t}{ }^{\prime}$ would be added to the final loss, $L_{\text {final }}$ by the proportionally updated distillation weight, $w^{\prime}$:
\vspace{-1mm}
\begin{equation}
L_{\text {final}}=L_{t}+w^{\prime} \cdot\left(L_{d v}+L_{d t}+L_{d t}^{\prime}\right)
\label{eq:ts}
\end{equation}

{\bf \noindent Confidence Weighted (CW) Distillation:} CLIP \cite{clip} benefits from a large amount of paired language-image training data. While this broad prior knowledge is likely helpful for VL tasks, the degree to which this knowledge is either helpful or potentially hurtful likely changes on an {\em instance level}. Given this, we design an approach to toggle the distillation objective depending on the relative confidence of the CLIP teacher and the specific student architecture. To do this, we define the ratio $r$ between maximum confidence scores of the CLIP teacher and base student model:
\vspace{-1mm}
\begin{equation}
r=T \frac{f_{argmax} \left(\sigma\left(L_{j}^{c}\right)\right)}{f_{argmax}\left(\sigma\left(L_{j}^{b}\right)\right)}
\end{equation}
\noindent where $L_{j}^{c}$ represents the logit vector from CLIP and $L_{j}^{b}$ from the base model. $T$ temperature is a hyperparameter value for adjusting the ratio. Finally, the new adaptive weight $w_{r}$ is defined as:
\vspace{-1mm}
\begin{equation}
w_{r}= \begin{cases}0 & \text {, if r } \leq 1 \\ w & \text {, if r }>1\end{cases}
\end{equation}

\noindent where $w_{r}$ replaces the distillation weights in Eq. \ref{eq:vl} and \ref{eq:ts}. When the ratio is above 1, CLIP is confident with its prediction, thus the distillation weight $w$ would be applied accordingly. Otherwise, the distillation value would be set to 0 to prevent from CLIP's interference.

{\bf \noindent Adaptive Finetuning (AF) with Contrastive Knowledge: } After large scale pretraining, previous work \cite{chen2020uniter} further trained V+L models with the same set of pretraining tasks, including Masked Language Modeling (MLM), Image-Text Matching (ITM), \textit{etc.} on downstream datasets. We denote those pretraining tasks as $L_{\text {pretraining }}$ (For details of $L_{\text {pretraining}}$, please refer to \cite{chen2020uniter}). Inspired by this, we further propose a two-stage finetuning strategy including Adaptive Finetuning with Contrastive Knowledge on downstream datasets. With our strategy, before the last-step finetuning with $L_{\text {final }}$, we finetune the base model with $L_{\text { AF }}$.
\vspace{-1mm}
\begin{equation}
L_{A F }=L_{\text {pretraining }}+w \cdot L_{d}
\end{equation}

\vspace{-5mm}
\subsubsection{Student Models}
\vspace{-2mm}
 Several recent top performing approaches for VL tasks are selected as students, including UNITER ~\cite{chen2020uniter}, VL-BERT ~\cite{Su2020VL-BERT}, and VILLA ~\cite{gan2020large}. All of these models represent variations of multi-modal architectures, using different portfolios of objective functions and datasets for pretraining, prior to finetuning for each respective VL task.


\vspace{-1mm}
\subsection{Direct Finetuning over CLIP}
\vspace{-1mm}
Inspired by prior work \cite{lu2021pretrained, hu2021lora}, we study various adapters composed of Multi-Layer Perceptron (MLP) layers or Transformer layers on top of CLIP. During finetuning for downstream tasks, we freeze the CLIP weights, and only finetune the adapters. Finally, we compare against concurrent work CLIP-VIL \cite{shengshen}, which more thoroughly finetunes CLIP models toward downstream VL tasks.



\vspace{-2mm}
\section{Datasets and Evaluations}
\vspace{-1mm}
We evaluate our methods on 3 commonly used vision-language task benchmarks: VCR, SNLI-VE, and VQA.

\subsection{Visual Commonsense Reasoning (VCR)}

The VCR benchmark presents images along with a paired question, a set of candidate answers, and a set of candidate rationales ~\cite{zellers2019vcr}. The dataset includes 290k questions, in reference to 110k unique visual scenes. The questions cover 7 categories of the type of inference needed, based on patterns in the questions. See supplement for a full list.

\vspace{-4mm}

\subsubsection{Data Availability}
\vspace{-1mm}

{\bf \noindent Zero-Shot:}
No training data is used. The pretrained model is directly employed to produce a matching between the image-question pair and a candidate answer. Answers are selected based on which produce the best matches, according to the model's matching measure.

{\bf \noindent Low-Shot:}
In the low-shot setting, we have 2 training set partitions of varying sizes: 100 examples per question category, totalling 700 pairs, or 0.3\% of the entire dataset, and  1,000 examples per category, totalling 7,000 pairs, or 3\%.

{\bf \noindent Semi-Supervised:}
In semi-supervised, we follow the same setting of Low-Shot for direct finetuning; however, the model has access to the remaining training data without groundtruth correct answer labels for distillation.

{\bf \noindent Fully Supervised:}
All question-answer pairs are used.
\vspace{-4mm}
\subsubsection{Evaluation Configurations}
\vspace{-1mm}
{\bf \noindent Standard Evaluation:} In this evaluation setting we follow the standard protocol in the benchmark.

{\bf \noindent Shortcut Mitigated (SM):}
We include a prior evaluation configuration ~\cite{debias} that focuses on mitigating shortcuts between question and answers. Shortcuts are shallow signals, cues models can learn to recognize that allow linking questions to correct answers without deep understanding of the content.  We refer to this as ``Shortcut Mitigated'' (SM).

{\bf \noindent Language Mitigation:}
Additionally, we introduce new evaluation protocols for the VCR dataset that further mitigate spurious language signals (Refer to supplement for more details):

{\em Explicit Mitigation (EM):}
Based on \cite{debias}, we employ a more generalized scope of defining shortcuts as frequently overlapped or co-occurred text between image-question and answer pairs. These apply for not only unigram but also $n$-gram scenarios. For each image-question pair, we extract tokens from text via tokenization and lemmatization. We also further extract object labels from the image  \cite{anderson2018bottom}. We then modify existing answer choices with the help from external knowledge database, \cite{miller1990introduction}. We replace highly-frequent instances of unigrams and $n$-grams with synonyms and hypernyms, thus eliminating these signals while simultaneously preserving the semantic and syntactic consistency of the text.


{\em Implicit Mitigation (IM):} Deep learning models can pick up trivial signals that humans cannot perceive \cite{arjovsky2019invariant}. For further mitigation, we trained a BERT language model to solve the VCR task using only the text. We then use this model to partition the test data into language-biased questions correctly answered by BERT with high confidence, and image-biased questions incorrectly answered by BERT. The cutoff confidence threshold, $\gamma$ is set at $90\%$.



\vspace{-2mm}
\subsection{Visual Entailment (SNLI-VE)}

The Stanford Natural Language Inference Visual Entailment (SNLI-VE) task ~\cite{xie2019visual} presents images as a premise, with paired hypothesis test. The goal is to predict whether the image entails or contradicts the hypothesis, or whether neither is the case (neutral). This is typically framed as a classification problem, since the predictions are fixed.
\vspace{-4mm}
\subsubsection{Data Availability}

{\bf \noindent Zero-Shot:}  The pretrained model is directly employed to produce the similarity between the image and the text premise. The challenge is determining what similarity values should constitute entailment, contradiction, or neither. To accomplish this, without finetuning the model, we perform k-means of the similarities on the validation set, with $k=3$, and use the resultant clusters as anchors for each output decision. Note that this approach, while a form of transductive learning on the validation set, uses no ground truth labels, and does not change any weights of the CLIP model.

{\bf \noindent Low-Shot:}
In the low-shot setting, we also have two settings: {\em 3,000} samples, with 1,000 random training datapoints for each class label, and {\em 30,000} samples, with 10,000 randomly sampled for each class label. They both correspond to $0.5\%$ and $5\%$ of the original training set.

{\bf \noindent Fully Supervised:}
All question-answer pairs are used for training or finetuning.
\vspace{-1mm}
\subsection{Visual Question Answering (VQA)}
\vspace{-1mm}
Different from VCR and VE, for every image-question pair in VQA \cite{VQA},  question-specific multiple choices are not provided. Instead, the global set of all possible answer choices for all the questions are provided (more than 3,000). The challenge is then to select the correct answer choice from this set for the given image-question pair. Comparing with VCR and VE, zero-shot on VQA is more challenging due to the large amount of answer choices. More importantly, all the answer choices for questions in VQA  consists of single words or short phrases. Thus, due to the semantic simplicity and ambiguity of the question and answer, a given question can potentially have more than one mapped answer choice.

\vspace{-5mm}
\subsubsection{Data Availability}
\vspace{-2mm}
{\bf \noindent Low-Shot:}
In the low-shot setting, we also have two settings: {\em 100} samples for training, and 1,000 random training datapoints.

{\bf \noindent Fully Supervised:}
All question-answer pairs are used for training or finetuning.

\begin{table*}[]
\centering
\begin{tabular}{l|ll|llll|llll}
\toprule
\textbf{Approach}         & \textbf{Patch} & \textbf{Method}  & \multicolumn{4}{l|}{\textbf{Standard   Evaluation}} & \multicolumn{4}{l}{\textbf{SM   Evaluation}} \\
\textbf{}            & \textbf{Size}              &                  & 0-shot & 700   & 7000  & Full                       & 0-shot         & 700           & 7000         & Full         \\ \midrule
\textbf{CLIP} & 32            & IQA              & 47.75  &       &       &                            & 51.56          &               &              &              \\
\textbf{0-shot}            & 32            & IA               & 49.22  &       &       &                            & 51.1           &               &              &              \\
\textbf{}            & 16            & IQA              & 50.27  &       &       &                            & {\bf 58.3}           &               &              &              \\
\textbf{}            & 16            & IA               & {\bf 54.82}  &       &       &                            & 53.82          &               &              &              \\ \hline
\textbf{CLIP}     & 32            & 1 Linear         &        & 30.13 & 30.39 & 31.86                      &                & 28.1          & 27.43        & 29.31        \\
\textbf{FineTune}            & 32            & 3 Linear         &        & 30.43 & 30.86 & 32.31                      &                & 28.73         & 28.32        & 29.94        \\
\textbf{}            & 32            & 1 Transformer    &        & 34.41 & 35.34 & 36.42                      &                & 33.35         & 34.02        & 35.48        \\ \hline
\textbf{VL-BERT}     & -             & Baseline         & 28.37  & 30.85 & 57.59 & \multicolumn{1}{r|}{76.02} & 25.24          & 26.42         & 54.53        & 72.21        \\
\textbf{Distillation}            & 32            & Naive VL               &        & 45.18 & 58.91 & 76.74                      &                & 42.99         & 55.66        & 73.5         \\
\textbf{}            & 32            & CLIP-TD (Ours)       &        & 46.74 & {\bf 59.98} & 77.24                      &                &  45.25          &  55.86        & 74.37        \\
\textbf{}            & 16            & CLIP-TD (Ours)      &        & {\bf 46.87} & 59.84 & {\bf 77.61}                      &                & {\bf 45.27}         & {\bf 55.88}        & {\bf 74.55}        \\ \hline
\textbf{UNITER}      & -             & B Baseline    & 29.42  & 31.31 & 57.02     & 74.23                      & 25.93          & 27.84         & 52.32        & 71.31        \\
\textbf{Distillation}           & 32            & B Naive VL          &        & 39.32 & 57.23 & 75.21                      &                & 40.23         & 54.24        & 72.12        \\
\textbf{}            & 32            & B CLIP-TD (Ours)  &        & 45.58 & 58.42 & 76.35                      &                & 44.65         & {\bf 54.95}        & 73.79        \\
                      & -              & L Baseline   & 29.78      & 31.43     & 58.24     & 76.67                      & 26.21              & 28.43             & 53.64           & 73.84        \\

\textbf{}            & 16            & L CLIP-TD (Ours) &        & {\bf 46.23}     & {\bf 59.47}     & {\bf 77.05}                      &                & {\bf 44.83}             & 54.93            & {\bf 74.23}        \\ \hline
\textbf{VILLA}            & -            & Baseline &        & 34.84     & 59.13     & 78.27                          &                & 29.41             & 54.15            & 75.43           \\
\textbf{Distillation}            & 16            & CLIP-TD (Ours) &        & \textbf{46.95}     & \textbf{60.13}     & \textbf{78.86}                          &                & \textbf{44.97}             & \textbf{55.22}            & \textbf{76.01}            \\ \hline
\textbf{ }            & 32            &  CLIP-ViL$_{p}$ \cite{shengshen} &        & 33.63     & 57.54     & 68.36                          &                & 30.41             & 55.42            & 66.83           \\ \bottomrule
\end{tabular}
\caption{VCR dataset results, including approaches for CLIP 0-shot, CLIP FineTune, and CLIP distillation to various student architectures. ``Patch Size'' refers to patch size of the CLIP teacher model (ViT-B16 vs. ViT-B32). Data sampling shown as number of image-question-answer triplets (0-shot to Full). For 0-shot: IQA = Image to Question+Answer cosine matching. IA = Image to answer cosine matching. For Distillation: B = Base, L = Large. SM = Shortcut Mitigated. ``Naive VL'' is naive distillation using vision and language class tokens. Results are based on our re-implementation of the students models with additional hyper-parameter tuning.}
\label{tab:vcr}
\end{table*}
\vspace{-3mm}
\begin{table*}[]
\centering
\begin{tabular}{l|ll|lll|lll}
\toprule
\textbf{Approach}        & \textbf{Patch} & \textbf{Method} & \multicolumn{3}{l|}{\textbf{Standard   Evaluation}} & \multicolumn{3}{l}{\textbf{SM Evaluation}} \\
\textbf{}        & \textbf{Size}              &                 & 700             & 7000            & Full           & 700                & 7000               & Full               \\ \midrule
\textbf{VL-BERT} & 32            & Naive Vision (V)          & 40.13           & 55.23           & 76.28          & 41.16              & 53.56              & 72.24              \\
\textbf{Baseline}        & 32            & Naive Language (L)        & 44.07           & 57.86           & 76.57          & 41.97              & 55.03              & 72.67              \\
\textbf{Distillation}        & 32            & Naive VL              & 45.18           & 58.91           & 76.74          & 42.99              & 55.66              & 73.5               \\
\textbf{}        & 32            & VL Random Token Selective (RTS)           & 40.21           & 57.24           & 75.34          & 42.02              & 55.43              & 77.23              \\ \hline
\textbf{Our}        & 32            & VL Token Selective (TS)           & 46.28           & 59.68           & 77.08          & 45.06              & 55.98              & 74.24              \\
\textbf{Distillation}                   & 32            & VL Confidence Weighted (CW)           & 46.54           & 58.93           & 76.84          & 45.17              & 55.76              & 74.23              \\
                 & 32            & VL Adaptive Finetune (AF)           & 46.02           & 59.23           & 76.93          & 45.26              & 55.87              & 74.54              \\
                 & 32            & VL AF + TS + CW (CLIP-TD)    & {\bf 46.74}           & {\bf 59.98}           & {\bf 77.24}          & {\bf 45.25}              & {\bf 55.86}              & {\bf 74.37}              \\ \bottomrule
\end{tabular}
\caption{VCR distillation ablation experiments using VL-BERT student model. ``Patch Size'' refers to patch size of the CLIP teacher model (ViT-B16 vs. ViT-B32). Training data subsampling shown under evaluation protocol. SM = Shortcut Mitigated.}
\label{tab:vcrablation}
\vspace{-3mm}
\end{table*}
\section{Results}
\subsection{Visual Commonsense Reasoning (VCR)}

Results on the VCR dataset for Q$\rightarrow$A, across several data availability and domain shifted constraints, are shown in Tab. \ref{tab:vcr}  (for additional metrics of Q2A and Q2AR, please see supplement). These results yield 5 key observations: 1) CLIP is capable of strong zero-shot results in VCR ``out-of-the-box'': over 58\% accuracy was attained on short-cut mitigated evaluation (a more difficult domain-shifted task), which is similar to some supervised approaches. 2) Our proposed CLIP-TD approach yields significant performance gains under low-shot data regimes across a range of student architectures: up to 52\% for VL-BERT, and 47.7\% for UNITER. 3) Our proposed CLIP-TD approach even benefits fully supervised tasks by 2.1\% for VL-BERT and 3.8\% for UNITER. 4) Our approach yields even higher gains under low-shot domain-shifted scenarios of shortcut mitigation (SM): up to 71.3\% for VL-BERT and 61\% for UNITER. 5) CLIP-TD also outperforms a prior proposed approach to leverage CLIP for VL tasks: CLIP-Vil \cite{shengshen}.

\begin{table}[]
\centering
\begin{tabular}{l|ll|ll}
\toprule
\textbf{Method} &  \multicolumn{2}{l|}{\textbf{Std. Evaluation}} & \multicolumn{2}{l}{\textbf{SM   Evaluation}} \\
                 &  700                      & 7000                     & 700                   & 7000                 \\ \midrule
 32, CLIP-TD         &  46.74                    & 59.98                    & 45.25                 & 55.86                \\
 32, Naïve VL (S)        &  51.32                    & 60.21                    & 48.24                 & 57.53                \\
 32, CLIP-TD (S)        &  51.93                    & 61.05                    & 48.33                 & 58.07                \\
 16, CLIP-TD (S)         & 52.43                    & 61.33                    & 48.71                 & 58.82               \\ \bottomrule
\end{tabular}
\caption{VCR semi-supervised using VL-BERT student model. Method denotes patch size of the CLIP teacher, distillation method, and whether Semi-supervised (S) or not. Training data sub-sampling shown under evaluation protocol. }
\label{tab:vcrsemi}
\vspace{-6mm}
\end{table}


\begin{center}
\begin{table*}[]
\centering
\begin{tabular}{lll|llll}
\toprule
\textbf{Approach} & \textbf{Patch} & \textbf{Method} &  \multicolumn{4}{l}{\textbf{Evaluation Mode}} \\
\textbf{}                & \textbf{Size}              &                 & Std    & SM     & IM    & EM    \\ \midrule
\textbf{VL-BERT}         &               & Baseline        & 76.02   & 72.21  & 70.64  & 68.23 \\
\textbf{Distillation}                & 16            & CLIP-TD (Ours)     & \textbf{77.61}  & \textbf{74.55}  & \textbf{73.67}  & \textbf{73.53} \\ \hline
\textbf{UNITER}          &               & B Baseline   & 74.23  & 70.54  & 69.93  & 69.42 \\
\textbf{Distillation}                & 16            & B CLIP-TD (Ours) & \textbf{76.35}  & \textbf{73.79}  & \textbf{71.35}  & \textbf{71.02} \\ \bottomrule
\end{tabular}
\caption{VCR Language Mitigation (ULM) results. ``Patch Size'' refers to patch size of the CLIP teacher model (ViT-B16 vs. ViT-B32). SM = Shortcut Mitigated. IM = Implicit Mitigation. EM = Explicit Mitigation}
\label{tab:vcrtextmitigation}
\vspace{-1mm}
\end{table*}
\end{center}

\vspace{-10mm}
Additionally, compared to training adapters on top of CLIP, or fine-tuning CLIP (concurrent work CLIP-ViL$_{p}$), using CLIP distillation to guide the learning of student models outperformed in all scenarios, demonstrating the effectiveness of this class of approaches.

CLIP-TD with VILLA delivers high performance on the public leaderboard (Q2A: 79.6\% QA2R: 82.9\% Q2AR: 66.2\%), achieving a new state-of-art of Q2A performance compared to other single models that are pretrained with image-text data only, as well as overall state-of-art comparable performance for QA2R and Q2AR.

As our approach is model agnostic, ensembles are possible. Combining multiple CLIP-TD approaches, further significant gains in performance yield up to \textbf{80.48\%} in the fully-sampled standard validation set.

As shown in Table \ref{tab:vcrablation}, we performed an ablation study of individual components of our CLIP-TD framework, evaluated with VL-BERT as the student model. These results demonstrate 3 key observations: 1) Language distillation, not vision (as studied in previous works), contributes the most performance improvement. However, distilling both vision and language perform better than either one alone. 2) Importantly, our token selective approach outperforms an approach that randomly selects the same number of tokens for distillation, demonstrating that the selective process is critical to maintain and improve performance. 3) CLIP-TD components each contributes a certain amount of performance improvement, and the benefits sum when all approaches are combined.

We also questioned how much performance gain in the low-shot settings might be accomplished with simple semi-supervised learning using CLIP distillation on data for which labels are withheld. Table \ref{tab:vcrsemi} shows the results studied using the VL-BERT model. In this setting, naive distillation in semi-supervised setting outperforms CLIP-TD without semi-supervision. However, if semi-supervision is used in combination with CLIP-TD, it leads to even further gains in comparison to naive distillation.

Table \ref{tab:vcrtextmitigation} shows results for both VL-BERT and UNITER baselines and CLIP-TD on our Unified Language Mitigation validation partitions. Under all conditions of language signal and shortcut mitigation, CLIP-TD produces results that significantly outperform the baselines, demonstrating an improved capability of leveraging true vision-language signals from the data.

Furthermore, we sought to better investigate how CLIP benefits performance of base student models. Following \cite{cao2020behind}, we measure Modality Importance (MI) of both modalities. This sums the attention weights across heads to understand how much each modality is weighted by the model. We compute the average MI values of all the heads for each layer on VL-BERT, both with and without CLIP-TD, trained on VCR. After distillation, we observe a 36\% reduction of MI difference between vision and text and a 10.6\% improvement in vision MI. For more details, see supplement.

\begin{figure}[t]
\begin{center}
\scriptsize
\scalebox{1}{
  \includegraphics[width=1\linewidth]{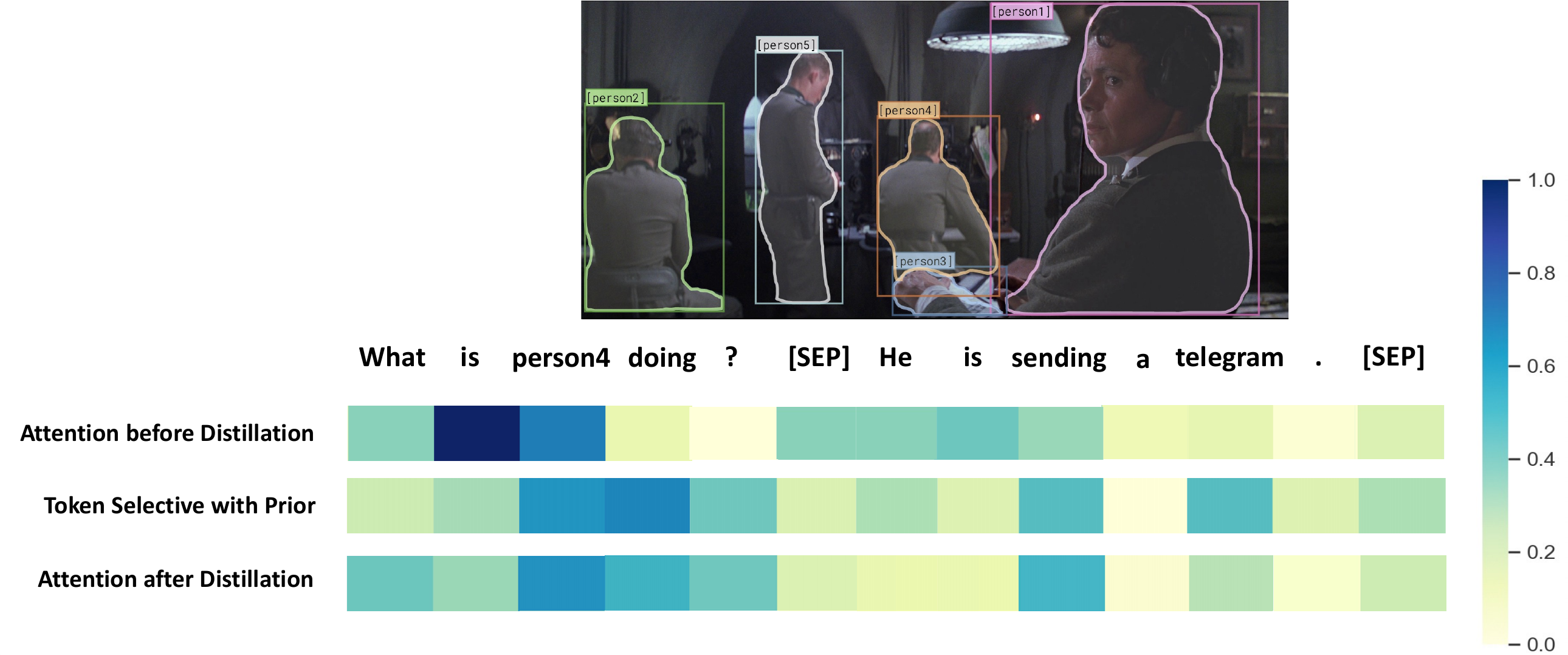}
}
\end{center}
 \vspace{-4mm}
  \caption{Comparison of attentions from VL-BERT before and after distillation, and Token Selective Module scores.}
\label{fig:sample_example}
\vspace{-5mm}
\end{figure}

Additionally, as V+L models are prone to learn trivial shortcuts from questions to correct answers \cite{kovaleva2019revealing, debias}, performance degrades when tested on datasets that mitigates these signals. We further seek to qualitatively better understand how our distillation approach might help improve the performance in this setting.  Fig. \ref{fig:sample_example}, shows token attention values from a VL-BERT model on an instance from the VCR dataset, before and after distillation, in addition to the token selection scores of our CLIP-TD approach. One can see that trivial tokens such as ``is'' have the highest attention values prior to distillation. By contrast, the token selection forces emphasis on more meaningful terms, such as ``sending,'' ``telegram,'' \textit{etc.}
\begin{table*}[h!]
\centering
\begin{tabular}{l|ll|llll|l}
\toprule
\textbf{Approach}     & \textbf{Patch} & \textbf{Method}        &\multicolumn{4}{c|}{\textbf{Validation}} & \textbf{Test} \\
\textbf{}     & \textbf{Size} & \textbf{}        & {0-shot}         & {3,000} & {30,000} & {Full} & {Full} \\\midrule
\textbf{CLIP 0-shot} & 32        & CLIP             & 46.83                   &                       &                      &                               &                       \\
\textbf{}            & 16        & CLIP             & 47.23                   &                       &                      &                               &                       \\ \hline
\textbf{VL-BERT}     &           & Baseline         & 33.31                   & 53.28                 & 62.31                & 74.66                         & 74.02                 \\
\textbf{Distillation}            & 16        & Naive VL               &                         & 56.02                 & 64.92                & 75.08                         & 74.93                 \\
\textbf{}            & 16        & CLIP-TD (Ours)        &                         & \textbf{56.78}                 & \textbf{65.37}                & \textbf{75.75}                         & \textbf{75.43}                 \\ \hline
\textbf{UNITER}      &           & B Baseline    & 32.42                   & 57.13                 & 65.52                & 78.53                         & 77.82                 \\
\textbf{Distillation}            & 16        & B Naive VL          &  & 58.24                 & 66.01                & 78.95                         & 78.67                 \\
\textbf{}            & 16        & B CLIP-TD (Ours)          &  & 58.27                 & 66.16                & 78.96                         & 78.83                 \\
\textbf{}            &           & L Baseline   &                         & 58.36                 & 66.23                & 79.02                         & 79.19                 \\
\textbf{}            & 16        & L Naive VL         &                         & 58.95                 & 67.74                & 80.08                         & 80.16                 \\
\textbf{}            & 16        & L CLIP-TD (Ours) &                         & \textbf{59.42}                 & \textbf{68.34}                & \textbf{80.14}                         & \textbf{80.23}                 \\ \hline
\textbf{VILLA}       &           & Baseline         & 36.72                       & 58.47                     & 67.16                    & 79.64                         & 79.32                 \\
\textbf{Distillation}            & 16        & CLIP-TD (Ours)       &                         & \textbf{59.65}                     & \textbf{68.43}                    & \textbf{80.67}                         & \textbf{80.31}                 \\ \hline
\textbf{Fine-Tuning}            &           & CLIP-ViL$_{p}$ \cite{shengshen}        &                         & 59.48                     & 68.32                    & 80.61                         & 80.2                  \\ \bottomrule
\end{tabular}
\caption{SNLI-VE Results. ``Patch Size'' refers to patch size of the CLIP teacher model (ViT-B16 vs. ViT-B32). Baselines represent the original methods without distillation. Training data subsampling shown under validation results. }
\label{tab:snlive}
\end{table*}

\begin{table*}[h!]
\centering
\begin{tabular}{c|cc|ccc|cc}
\toprule
\textbf{Approach}     & \textbf{Patch} & \textbf{Method}                 & \multicolumn{3}{c|}{\textbf{Validation}} & \textbf{Test-Val} & \textbf{Test-Std} \\
                      & \textbf{Size}  &                                 & 100       & 1,000    & Full     &                   &                   \\ \midrule
\textbf{VL-BERT}      &                & Baseline                        & 35.33     & 63.29    & 69.05    & 71.79             & 72.22             \\
\textbf{Distillation} & 16             & Naive VL                        & 36.83     & 64.43    & 70.22    &                   &                   \\
                      & 16             & CLIP-TD (Ours)                  & 37.12     & 65.71    & 71.42    &                   &                   \\ \hline
\textbf{UNITER}       &                & L Baseline                      & 36.46     & 64.43    & 71.26    & 73.82             & 74.02             \\
\textbf{Distillation} & 16             & Naive VL                        & 38.95     & 65.17    & 71.56    &                   &                   \\
                      & 16             & L CLIP-TD (Ours)                & 39.75     & 65.93    & 71.94    &                   &                   \\ \hline
\textbf{VILLA}        &                & Baseline                        & 37.18     & 65.75    & 72.11    & 74.69             & 74.87             \\
\textbf{Distillation} & 16             & CLIP-TD (Ours)                  & \textbf{40.16}     & \textbf{66.93}    & 73.02    & 75.81             & 76.04             \\ \hline
\textbf{Fine-Tuning}  &                & CLIP-ViL$_{p}$ \cite{shengshen} & 39.01     & 66.84    & \textbf{73.91}    & \textbf{76.48}             & \textbf{76.70}              \\ \bottomrule
\end{tabular}
\caption{VQA Results. ``Patch Size'' refers to patch size of the CLIP teacher model (ViT-B16 vs. ViT-B32). L = Large model version of UNITER. Baselines represent the original methods without distillation. Training data subsampling shown under validation results. Test results were accomplished using full training set.}
\label{tab:vqa}
\vspace{-4mm}
\end{table*}

\vspace{-1mm}
\subsection{Visual Entailment (SNLI-VE)}
\vspace{-1mm}
Results on SNLI-VE are shown in Table \ref{tab:snlive}, revealing 3 key observations: 1) CLIP can achieve up to 40.02 \% zero-shot accuracy on val set of SNLI-VE ``out-of-the-box'', which is significantly higher than baseline methods like VL-BERT and UNITER. 2) CLIP-TD continues to provide accuracy improvements in all settings of low-shot and fully-supervised conditions, using all student base models, versus both baseline with no distillation, and naive distillation. 3) CLIP-TD continues to outperform concurrent work CLIP-ViL$_{p}$ under all evaluated conditions.

With CLIP-TD, base models like VL-BERT or UNITER Base can realize improvement of 6.5 \% with only 3,000 samples and 4.9 \% with 30,000. In addition, CLIP-TD consistently brings improvement to all base models, demonstrating that this approach is model agnostic. In conjunction with VILLA, CLIP-TD can push the performance to 80.67 \% on validation and 80.31 \% on test, both of which outperform CLIP-VIL \cite{shengshen}.



\vspace{-1mm}
\subsection{Visual Question Answering (VQA)}
\vspace{-1mm}

Table \ref{tab:vqa} shows our results from VQA. The key observations from these experiments are as follows: 1) CLIP-TD yields performance improvement against baseline and naive distillation under all conditions. 2) Under low-shot conditions, CLIP-TD can outperform finetuning approach CLIP-ViL$_{p}$  3) Under the full-shot setting, CLIP-TD can assist the baseline method, VILLA to reach performance of 73.02 on validation, which is comparable to the performance from CLIP-VIL$_{p}$. Further experimentation leveraging the model-agnostic CLIP-TD to create ensembles achieves \textbf{73.93 \%} on local validation, outperforming CLIP-VIL$_{p}$.




\vspace{-3mm}
\section{Conclusion}
In this work, we seek to better understand the utility CLIP presents for VL tasks. To accomplish this, we present two key contributions: 1) We present a new VL task protocol over VCR, SNLI-VE, and VQA that covers zero-shot, low-shot, semi-supervised, fully-supervised, and domain shifted settings. 2) We evaluate a newly proposed approach, CLIP-TD, to perform targeted distillation from CLIP to a variety of high performing base student models, including VL-BERT, UNITER, and VILLA. Our results demonstrate significant improvements in performance in all datasets, tasks, and settings, compared with baselines without CLIP distillation, and naive distillation methods. Compared to concurrent fine-tuning work CLIP-ViL, our approach outperforms under all low-shot conditions and most fully-supervised conditions, except for fully-supervised VQA, where the performance is comparable. Because our approach is agnostic to the base student model, ensembles of many student models can yield superior results under all conditions studied. We believe our results can be helpful for the community to know how to best leverage CLIP to potentially boost the performance of future works on VL tasks.

\paragraph{Acknowledgement}
Thanks to Liunian Harold Li for his help in the implementation of CLIP-ViL and feedbacks of the idea.

{\small
\bibliographystyle{ieee_fullname}
\bibliography{egbib}
}

\clearpage

\pagebreak

\pagebreak[4]


\section{Supplementary Material}


\begin{table*}[th!]
\centering
\scalebox{1}{
\begin{tabular}{l|clccc}
\toprule
\multicolumn{1}{c|}{\textbf{Approach}}                                   & \textbf{Patch} & \multicolumn{1}{c}{\textbf{Method}} & \multicolumn{3}{c}{\textbf{Standard Evaluation}} \\

\multicolumn{1}{c|}{}                                       & \textbf{Size}  & \multicolumn{1}{c}{}                & Q2A              & QA2R            & Q2AR            \\ \midrule
\multicolumn{1}{c|}{\textbf{Zero-shot}}                                     & 16             & IA(R)                               & 54.82            & 48.58           & 26.63           \\ \hline
\textbf{VL-BERT}                                                         & -              & Baseline                            & 76.02            & 78.31           & 59.53           \\
\textbf{Distillation}                                                    & 32             & Naive VL                            & 76.74            & 78.64           & 60.35           \\
\textbf{}                                                                & 32             & CLIP-TD (Ours)                      & 77.24            & 79.02           & 61.04           \\
\textbf{}                                                                & 16             & CLIP-TD (Ours)                      & \textbf{77.61}   & \textbf{79.25}  & \textbf{61.51}  \\ \hline
\textbf{UNITER}                                                          & -              & B Baseline                          & 74.23            & 76.99           & 57.15           \\
\textbf{Distillation}                                                    & 32             & B Naive VL                          & 75.21            & 77.59           & 58.36           \\
\textbf{}                                                                & 32             & B CLIP-TD (Ours)                    & 76.35            & 77.84           & 59.43           \\
\textbf{}                                                                & -              & L Baseline                          & 76.67            & 79.98           & 61.32           \\
\textbf{}                                                                & 16             & L CLIP-TD (Ours)                    & \textbf{77.05}   & \textbf{80.57}  & \textbf{62.08}  \\ \hline
\textbf{VILLA}                                                           & -              & Baseline                            & 78.27            & 82.33           & 64.44           \\
\textbf{Distillation}                                                    & 16             & CLIP-TD (Ours)                      & \textbf{78.86}   & \textbf{82.57}  & \textbf{65.11}  \\ \hline
\textbf{\begin{tabular}[c]{@{}l@{}}Ensemble\\ Distillation\end{tabular}} & 16             & CLIP-TD (Ours)                      & \textbf{80.48}   & \textbf{82.68}  & \textbf{66.54}  \\ \hline
                                                                         & 32             & CLIP-ViL$_{p}$ \cite{shengshen}     & 68.36            & 71.40           & 48.81           \\ \bottomrule
\end{tabular}
}
\caption{Complete VCR Evaluation Results}
\label{vcr_full}
\end{table*}

Section \ref{sec:vcr} covers additional details regarding our evaluations on the VCR dataset, including our public leaderboard results (\ref{sec:vcr-publicresults}), additional metrics for internal validation results (\ref{sec:vcr-valresults}), information about question sub-types (\ref{sec:vcr-subtypes}), details about language mitigation (\ref{sec:vcr-lm}), and further analysis of model behavior before and after distillation (\ref{sec:vcr-analysis}).

Section \ref{sec:trainingdetails} provides additional details regarding training code configurations (\ref{sec:trainingdetails-baselines}) and parameters in all of our experiments (\ref{sec:trainingdetails-implementation}).

Section \ref{sec:baselinemethod} covers additional details regarding our baselines for naive distillation (\ref{sec:baselinemethod-nkd}) and adapters over CLIP (\ref{sec:baselinemethod-adapters}).

\section{VCR Evaluation}
\label{sec:vcr}
\subsection{Public Leaderboard Results}
\label{sec:vcr-publicresults}

We submitted our single model test prediction results to the VCR public leaderboard, which is currently listed as the 9th entry overall (inlcuding ensemble models) and achieves state-of-art performance on VCR compared to other single models that are pretrained with image-text data only. The entry is displayed under name, CLIP-TD with test result: Q2A $79.6 \%$ QA2R $82.9 \%$ Q2AR $66.2 \%$. Meanwhile, we are also submitting our ensemble models' prediction results to the leaderboard, as shown in the second last row of Tab. \ref{vcr_full}.

\subsection{Standard Validation Results}
\label{sec:vcr-valresults}

As in Tab. \ref{vcr_full}, we include comparison of our methods on top of top-performing base models for all three evaluation metrics: Q2A, QA2R and Q2AR. Our method consistently improves on top of the base models across all metrics.

\subsection{VCR Dataset Question Sub-Types}
\label{sec:vcr-subtypes}

According to \cite{zellers2019vcr}, among VCR questions, 38\% fall into explanation (why, how come, \textit{etc.}), 24\% activity (doing, looking, event, \textit{etc.}), 13\% temporal (happened, before, after, \textit{etc.}), 8\% mental (feeling, thinking, \textit{etc.}), 7\% role (relations, occupations, \textit{etc.}), 5\% scene (where, near, \textit{etc.}), and 5\% hypothetical (if, would, could, \textit{etc.}). Details can be referred to \cite{zellers2019vcr}.

\subsection{Language Mitigation Details}
\label{sec:vcr-lm}

Inspired by \cite{debias} which only focuses on modiying pronouns withing the original text in VCR, we introduce new evaluation protocols for the VCR dataset that further mitigate spurious language signals in a much more general perspective.

\begin{table*}[]
\centering
\scalebox{0.7}{
\begin{tabular}{l|clll|ccc|ccc}
\hline
\multicolumn{1}{c|}{\textbf{Model}} & \textbf{Patch} & \multicolumn{1}{c}{\textbf{Method}}                & \multicolumn{1}{c}{\textbf{Distillation}} & \multicolumn{1}{c|}{\textbf{Token \#}} & \multicolumn{3}{c|}{\textbf{Standard   Evaluation}} & \multicolumn{3}{c}{\textbf{SM Evaluation}}       \\
\multicolumn{1}{c|}{}               & \textbf{Size}  & \multicolumn{1}{c}{}                               & \multicolumn{1}{c}{\textbf{Weight}}       & \multicolumn{1}{c|}{}                  & 700             & 7000            & Full            & 700            & 7000           & Full           \\ \hline
\textbf{}                           &                &                                                    &                                           &                                        & 30.85           & 57.59           & 76.02           & 26.42          & 55.03          & 72.21          \\ \cline{2-11}
                                    & 32             & Naive VL                                           & \multicolumn{1}{c}{1.0}                   &                                        & 35.39           & 58.57           & 76.4            & 32.2           & 55.05          & 72.65          \\
\textbf{}                           & 32             & Naive VL                                           & \multicolumn{1}{c}{0.5}                   &                                        & 37.20           & 58.96           & 76.42           & 35.31          & 55.17          & 73.10          \\
                                    & 32             & Naive VL                                           & \multicolumn{1}{c}{0.1}                   &                                        & 40.47           & 58.72           & 76.62           & 37.64          & 55.32          & 73.20          \\
                                    & 32             & Naive VL                                           & \multicolumn{1}{c}{0.05}                  &                                        & 45.18           & 58.91           & 76.74           & 42.99          & 55.66          & 73.50          \\
\textbf{}                           & 32             & Naive VL                                           & \multicolumn{1}{c}{0.01}                  &                                        & 42.31           & 58.65           & 76.23           & 41.34          & 55.23          & 73.03          \\ \cline{2-11}
VL-BERT                             & 32             & VL Random Token Selective (RTS)                    &                                           & \multicolumn{1}{c|}{2}                 & 40.21           & 57.24           & 75.34           & 42.02          & 55.43          & 77.23          \\
                                    & 32             & VL Token Selective (TS)                            &                                           & \multicolumn{1}{c|}{1}                 & 45.24           & 58.88           & 76.71           & 43.12          & 55.40          & 73.51          \\
                                    & 32             & VL Token Selective (TS)                            &                                           & \multicolumn{1}{c|}{2}                 & 46.28           & 59.68           & 77.08           & 45.06          & 55.98          & 74.24          \\
                                    & 32             & VL Token Selective (TS)                            &                                           & \multicolumn{1}{c|}{3}                 & 45.84           & 59.07           & 76.87           & 44.22          & 55.49          & 73.73          \\ \cline{2-11}
\textbf{}                           & 32             & VL Confidence Weighted (CW) w/ correct answer only &                                           &                                        & 46.14           & 58.27           & 76.66           & 44.83          & 55.64          & 73.79          \\
\textbf{}                           & 32             & VL Confidence Weighted (CW) w/ all answer choices  &                                           &                                        & 46.54           & 58.93           & 76.84           & 45.17          & 55.76          & 74.23          \\ \cline{2-11}
                                    & 32             & 2nd-stage Pretraining                              &                                           &                                        & 38.32           & 58.13           & 76.38           & 36.42          & 55.07          & 72.87          \\
                                    & 32             & VL Adaptive Finetune (AF)                          &                                           &                                        & 46.02           & 59.23           & 76.93           & 45.26          & 55.87          & 74.54          \\ \cline{2-11}
                                    & 32             & VL AF + TS + CW (CLIP-TD)                          &                                           & \multicolumn{1}{c|}{2}                 & \textbf{46.74}  & \textbf{59.98}  & \textbf{77.24}  & \textbf{45.25} & \textbf{55.86} & \textbf{74.37} \\ \hline
\end{tabular}
}
\caption{VCR distillation ablation experiments using VL-BERT student model. ``Patch Size'' refers to patch size of the CLIP teacher model (ViT-B16 vs. ViT-B32). Training data subsampling shown under evaluation protocol. SM = Shortcut Mitigated.}
\label{ablation}
\end{table*}

{\em Explicit Mitigation (EM):}
Based on \cite{debias}, we employ a more generalized scope of defining shortcuts as frequently overlapped or co-occurred text between image-question and answer pairs. These apply for not only unigram but also $n$-gram scenarios. For each image-question pair, we extract tokens from text via tokenization and lemmatization. We also further extract object labels from the image  \cite{anderson2018bottom}. As in Tab. \ref{vcr_bias}, the top two rows represent statistics based on the full standard validation set (without Implicit Mitigation filtering). It is obvious to observe that the average number of overlapped tokens between the question and the correct answer choice is much higher than the incorrect answer choice (We also average among three incorrect answer choices per question). In around 44.44 $\%$ samples of the full validation set, correct answer choices have more overlapped tokens than the incorrect answer choices. This explicit syntactic advantage clearly favors the correct answer choice. Despite it is trivial, V+L models may easily pick it up and it is hard to eliminate due to problems like annotator bias/laziness, \textit{etc.} After further filtering with Implicit Mitigation (we remove all the samples where the language-only model cam answer correctly with a very high confidence), this advantage/syntactic bias for the correct answer becomes even more obvious. This drastic increase further proves our suspicion that existing models are inevitably influenced by the bias. We then modify existing answer choices with the help from external knowledge database, \cite{miller1990introduction}. We replace highly-frequent instances of unigrams and $n$-grams with synonyms and hypernyms, thus eliminating these signals while simultaneously preserving the semantic and syntactic consistency of the text.

{\em Implicit Mitigation (IM):} Deep learning models can pick up trivial signals that humans cannot perceive \cite{arjovsky2019invariant}. For further mitigation, we trained a BERT language model to solve the VCR task using only the text. We then use this model to partition the test data into language-biased questions correctly answered by BERT with high confidence, and image-biased questions incorrectly answered by BERT. The cutoff confidence threshold, $\gamma$ is set at $90\%$.

\subsection{Further Analysis of CLIP Distillation}
\label{sec:vcr-analysis}

\begin{figure}[th!]
\begin{center}
\scriptsize
 \includegraphics[]{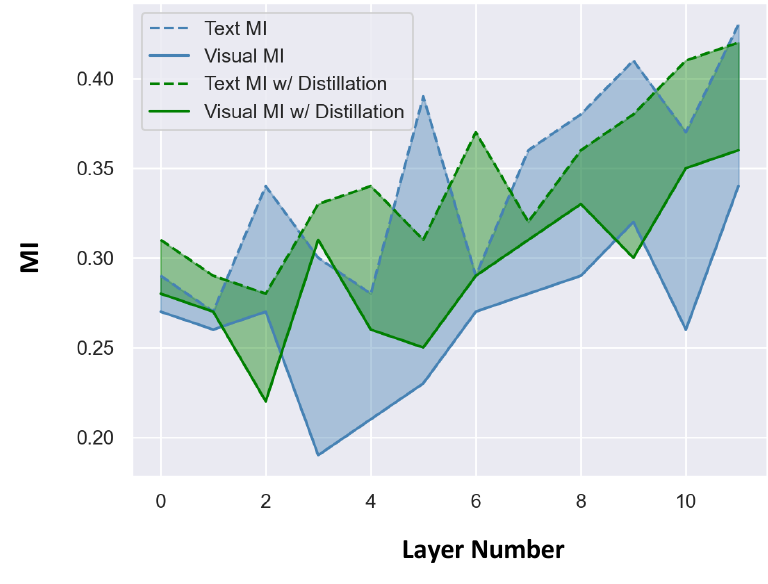}
\end{center}
 \caption{Average Modality Importance (MI) values for each layer of VL-BERT with CLIP-TD (green) and baseline (blue). Shaded areas represent difference between visual and textual modalities.}
\label{fig:mi_line}
\end{figure}

\begin{figure}[t]
\begin{center}
\scriptsize
 \includegraphics[width=1\linewidth]{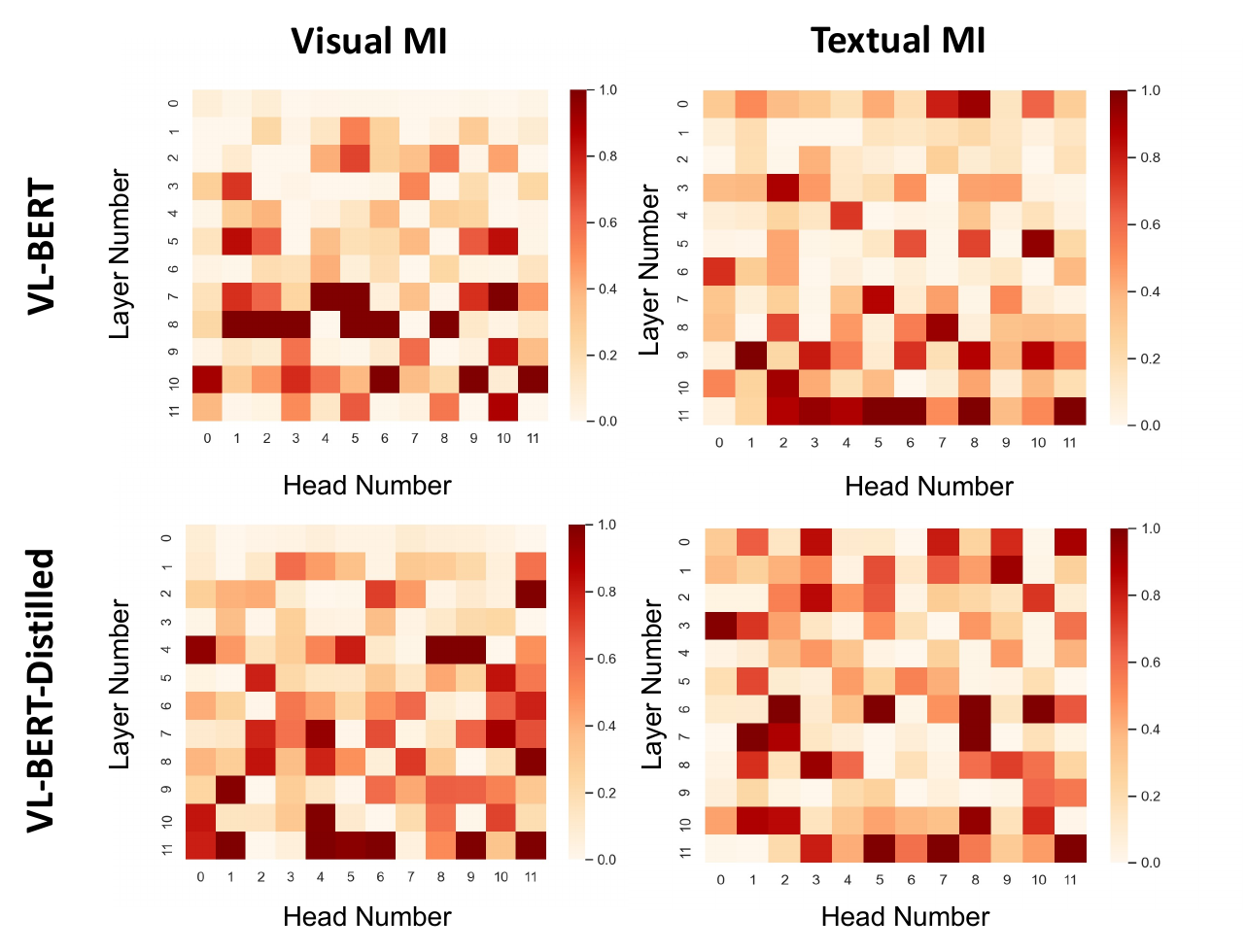}
\end{center}
 \caption{Heatmap of MI values for both visual and textual modality. Each cell represents a combination between 12 heads and 12 layers in VL-BERT Base model. The top row corresponds to the original VL-BERT Base model and the bottom row corresponds to VL-BERT Base with CLIP-TD. Index starts with 0.}
\label{fig:mi_heat}
\vspace{-4mm}
\end{figure}

Based on our results, we have seen the impact on end task performance of distilling knowledge from CLIP's dual independent encoders into student V+L models. However, a key question still remains about whether the improvement solely comes from large amount of contrastive pretraining or also from other perspectives?


Following \cite{cao2020behind}, we measure the Modality Importance (MI) of both visual modality and textual modality. This approach sums the attention weights across heads of each modality to understand how much each modality is weighted by the model. Fig. \ref{fig:mi_line} shows the average the MI values of all the heads for each layer on VL-BERT, both with and without CLIP-TD, trained on the VCR dataset. One can clearly observe that prior to distillation, the model more heavily weights the text modality as being important to correctly choosing answers. After distillation, both vision and text modalities are more equally considered. This may also explain why the model yields such impressive performance improvements in low-shot and domain shifted scenarios.

Fig. \ref{fig:mi_heat}, we plot the MI values of all the heads across 12 layers in VL-BERT Base and VL-BERT Base with CLIP-TD. It is obvious that, at the last layer, the textual MI heatmap on the right is denser than the visual MI heatmap on the left. This shows a common flaw from existing V+L models that they heavily rely on the textual information than the visual part indicating the shallow understanding of the visual scene in downstream tasks. However, in the bottom row, the difference between the left and right heatmaps is much smaller and the visual MI heatmap at the bottom is also clearly more denser than the one at the top. This could be further verified by

\begin{table}[]
\scalebox{0.75}{
\centering
\begin{tabular}{cccccc}
\hline
\multicolumn{1}{l|}{\textbf{Category}} & \multicolumn{1}{c|}{\textbf{IM}} & \multicolumn{2}{c|}{\textbf{Overlap w/ Question}}                                                                                                                                           & \multicolumn{2}{c}{\textbf{Overlap w/ Image}}                                                                                                                          \\ \cline{3-6}
\multicolumn{1}{l|}{}                  & \multicolumn{1}{c|}{}            & \multicolumn{1}{c|}{\textbf{\begin{tabular}[c]{@{}c@{}}Avg.\\ Token No.\end{tabular}}} & \multicolumn{1}{c|}{\textbf{\begin{tabular}[c]{@{}c@{}}Percentage of \\ More Overlaps\end{tabular}}} & \multicolumn{1}{c|}{\textbf{\begin{tabular}[c]{@{}c@{}}Avg.\\ Token No.\end{tabular}}} & \textbf{\begin{tabular}[c]{@{}c@{}}Percentage of \\ More Overlaps\end{tabular}} \\ \hline
Incorrect                              & N                                & 1.99                                                                                 & 38.16$\%$                                                                                            & 0.95                                                                                 & 32.23$\%$                                                                       \\
Correct                                & N                                & 2.21                                                                                 & 44.44$\%$                                                                                            & 0.97                                                                                 & 45.23$\%$                                                                       \\ \hline
Incorrect                              & Y                                & 1.65                                                                                 & 12.11$\%$                                                                                            & 0.83                                                                                 & 28.84$\%$                                                                       \\
Correct                                & Y                                & 3.14                                                                                 & 77.81$\%$                                                                                            & 0.98                                                                                 & 48.16$\%$                                                                       \\ \hline
\end{tabular}}
\caption{VCR dataset text bias measurements. ``Answer Category'' column groups answers according to correct and incorrect answers. ``IM'', or Implicit Mitigation, groups answers according to whether or not a simple BERT language model was able to select the correct answer. ``Token Overlap'' denotes the overlap between question and answer sentences for the given partition. }
\label{vcr_bias}
\end{table}

\begin{center}
\begin{table*}[]
\centering
\begin{tabular}{lll|llll}
\toprule
\textbf{Approach} & \textbf{Patch} & \textbf{Method} &  \multicolumn{4}{l}{\textbf{Evaluation Mode}} \\
\textbf{}                & \textbf{Size}              &                 & Std    & SM     & IM    & EM    \\ \midrule
\textbf{VL-BERT}         &               & Baseline        & 76.02   & 72.21  & 70.64  & 68.23 \\
\textbf{Distillation}                & 16            & CLIP-TD (Ours)     & \textbf{77.61}  & \textbf{74.55}  & \textbf{73.67}  & \textbf{73.53} \\ \hline
\textbf{UNITER}          &               & B Baseline   & 74.23  & 70.54  & 69.93  & 69.42 \\
\textbf{Distillation}                & 16            & B CLIP-TD (Ours) & \textbf{76.35}  & \textbf{73.79}  & \textbf{71.35}  & \textbf{71.02} \\ \bottomrule
\end{tabular}
\caption{VCR Language Mitigation (ULM) results. ``Patch Size'' refers to patch size of the CLIP teacher model (ViT-B16 vs. ViT-B32). SM = Shortcut Mitigated. IM = Implicit Mitigation. EM = Explicit Mitigation}
\label{tab:vcrtextmitigation}
\vspace{-1mm}
\end{table*}
\end{center}

\begin{table}[]
\scalebox{0.6}{
\centering
\begin{tabular}{cccccccccc}
\hline
\textbf{Model}                                                                                      & \textbf{Adapter}                                           & \textbf{Layer}       & \textbf{Fusion}      & \multicolumn{3}{c}{\textbf{Standard Evaluation}} & \multicolumn{3}{c}{\textbf{SM Evaluation}} \\ \hline
\multicolumn{1}{c|}{\multirow{6}{*}{\begin{tabular}[c]{@{}c@{}}CLIP\\ \\   (VIT-B32)\end{tabular}}} & \multicolumn{1}{l}{}                                       & \multicolumn{1}{l}{} & \multicolumn{1}{l}{} & 700            & 7000           & Full           & 700          & 7000         & Full         \\ \cline{2-10}
\multicolumn{1}{c|}{}                                                                               & \multirow{4}{*}{MLP}                                       & 1                    & Concat.              & 30.13          & 30.39          & 31.86          & 28.1         & 27.43        & 29.31        \\
\multicolumn{1}{c|}{}                                                                               &                                                            & 1                    & Cos.                 & 28.64          & 29.31          & 30.42          & 27.32        & 27.04        & 29.03        \\
\multicolumn{1}{c|}{}                                                                               &                                                            & 3                    & Concat.              & 30.43          & 30.86          & 32.31          & 28.73        & 28.32        & 29.94        \\
\multicolumn{1}{c|}{}                                                                               &                                                            & 3                    & Cos.                 & 29.65          & 30.23          & 30.84          & 28.85        & 28.08        & 29.64        \\ \cline{2-10}
\multicolumn{1}{c|}{}                                                                               & \begin{tabular}[c]{@{}c@{}}Attention \\ Layer\end{tabular} & 1                    & Concat.              & 34.41          & 35.34          & 36.42          & 33.35        & 34.02        & 35.48        \\ \hline
\end{tabular}
}
\caption{Adapters over CLIP. Concat. represents concatenation of visual and language features. Cos. represents
cosine similarity between visual and language features.
}
\label{adapter}
\end{table}

\section{Training Details}
\label{sec:trainingdetails}
\subsection{Baselines}
\label{sec:trainingdetails-baselines}

All the baseline experiment results with models including VL-BERT, UNITER Base, UNITER Large, VILLA and CLIP-ViL$_{p}$ are based on code provided by the authors, which we modified to include our distillation methods. CLIP-ViL$_{p}$ did not originally evaluate on VCR in their pre-print. However, we evaluated it on the VCR dataset.

In our paper, the performance of a base model, VL-BERT is slightly higher than the listed performance in the original paper. This increase is likely due to our hyper-parameter tuning.

\subsection{Implementation Details}
\label{sec:trainingdetails-implementation}

\textbf{- Token Selective (TS) Distillation: } Our ablation experiments with TS distillation show that when the number of tokens selected is 2, the highest performance is obtained, shown in Tab. \ref{ablation}.

\textbf{- Confidence Weighted (CW) Distillation:} With Confidence Weighted (CW) knowledge distillation method, during our experiment, we find out that the performance would achieve the optimal gain when the distillation is conducted across all four question-answer pairs instead of question-correct-answer pair only, as in Tab. \ref{ablation}.

\textbf{- Adaptive Finetuning (AF) with Contrastive Knowledge:} Before the last-step finetuning for the target downstream tasks, we conduct Adaptive Finetuning to adapt the model to the downstream domain. During the Adaptive Finetuning, the model is trained on the full training set. The loss contains the construction loss from the same set of pretraining tasks like Masked Language Modeling (MLM), Image-Text Matching (ITM), \textit{etc.} as in \cite{chen2020uniter}. Additionally, we also include Naive Knowledge Distillation thus the final loss also includes the knowledge distillation loss besides the construction loss.

\textbf{- VL-BERT: } We train our model for 30 epochs with warm-up steps of 1000, SGD optimizer. Initial learning rate is $7.0e-5$ and  decays by 0.1 at the 14th, 18th and 26th epoch. The gradient accumulation steps is set to be 4 on 8 NVIDIA V100 GPUs (32GB VRAM). The total number of layers for VL-BERT is 24 VL-BERT$_{Large}$.

\textbf{- UNITER: }  Started with warm up steps of 800, the model is trained with total steps of 8000. With AdamW optimizer, the intial learning rate is set to be $6e-05$ with weight decay of 0.01 and batch size of 4000. The gradient accumulation steps is set to be 5 on 4 NVIDIA TITAN RTX GPUs (24GB VRAM).

\textbf{- VILLA: } Warm up steps is set to be 1000 and total training steps is 10000. The intial learning rate is $6e-05$ with weight decay of 0.01 and AdamW optimizer. The training batch size is 1250. The gradient accumulation steps is set to be 8 on 8 NVIDIA TITAN RTX GPUs (24GB VRAM).

\textbf{- CLIP-ViL$_{p}$: } The model is trained for 20 epochs with batch size of 24. The optimizer is AdamW with a peak learning rate of 5 x $10^{-5}$.

\section{Baseline Method}
\label{sec:baselinemethod}
\subsection{Naive Knowledge Distillation}
\label{sec:baselinemethod-nkd}

\begin{figure}[]
\centering
\includegraphics[width=1\linewidth]{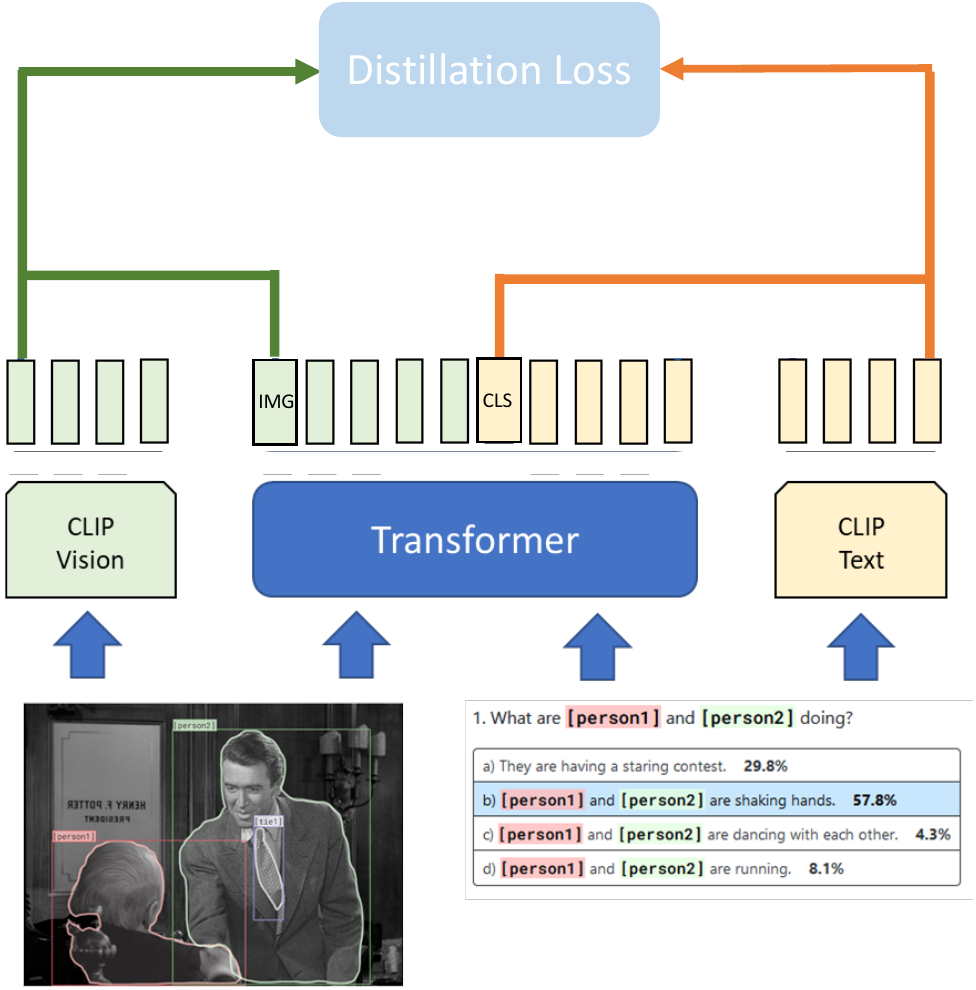}
\caption{The proposed Naive Knowledge Distillation method. For a given task with an established base model as the student. The l1 loss is calculated respectively for both the visual and text representation features between the teacher and the student.}
\vspace{-5mm}
\label{fig:naive_kd}
\end{figure}

As illustrated in Fig. \ref{fig:naive_kd} and the top rows of Tab. \ref{ablation}, we experiment with a wide spectrum of disllation weight and realize that the performance is best optimized when the weight is set to be 0.05.
\subsection{Adapter over CLIP}
\label{sec:baselinemethod-adapters}
For more comprehensively exploring different options to utilize CLIP's pretrained knowledge for downstream Vision-Language tasks, we also experimented to directly add adapters on top of CLIP. As listed in Tab. \ref{adapter}, we eseentially have experimented with adding either MLP or attention layers on top of CLIP. However, due to the large gap between the size of finetuning data and CLIP's original pretraining data, the adapters' limited capacity fail to adapt the CLIP's preatrained knowledge for the downstream tasks efficiently.

\end{document}